\documentclass[lettersize]{IEEEtran}

\usepackage{amsmath,amsfonts}
\usepackage{algorithm}
\usepackage{algpseudocode}
\usepackage{array}
\usepackage[caption=false,font=normalsize,labelfont=sf,textfont=sf]{subfig}
\usepackage{textcomp}
\usepackage{stfloats}
\usepackage{url}
\usepackage{balance}
\usepackage{verbatim}
\usepackage{xcolor}
\usepackage{graphicx}
\usepackage{cite}
\usepackage{longtable}
\usepackage{multirow}
\usepackage{booktabs}

\usepackage[utf8]{inputenc}
\usepackage{siunitx}
\usepackage{hyperref}
\usepackage{amsmath}
\usepackage{amssymb}
\usepackage{authblk}

\title{Distributed Deep Learning for Medical Image Denoising with Data Obfuscation}

\author[1]{Sulaimon Oyeniyi Adebayo}
\author[1, 2]{Ayaz H. Khan}

\affil[1]{Computer Engineering Department, King Fahd University of Petroleum and Minerals, Dhahran, 31261, Saudi Arabia}
\affil[2]{SDAIA-KFUPM Joint Research Center for Artificial Intelligence, King Fahd University of Petroleum and Minerals, Dhahran 31261, Saudi Arabia}

\begin{document}

\maketitle

\begin{abstract}
Medical image denoising is essential for improving image qualitywhile minimizing the exposure of sensitive information, particularly when working with large-scale clinical datasets. This study explores distributed deep learning for denoising chest X-ray images from the NIH Chest X-ray14 dataset, using additive Gaussian noise as a lightweight obfuscation technique. We implement and evaluate U-Net and U-Net++ architectures under single-GPU, standard multi-GPU (DataParallel), and optimized multi-GPU training configurations using PyTorch’s DistributedDataParallel (DDP) and Automatic Mixed Precision (AMP). Our results show that U-Net++ consistently delivers superior denoising performance, achieving competitive Peak Signal to Noise Ratio (PSNR) and Structured Similarity Index Method (SSIM) scores, though with less performance in Learned Perceptual Image Patch Similarity (LPIPS) compared to U-Net under low and moderate noise levels. This indicates U-Net++'s enhanced structural fidelity and low perceptual similarity. Meanwhile, our optimized training pipeline reduces training time by over 60\% for both models compared to single-GPU training, and outperforms standard DataParallel by over 40\%, with only a minor accuracy drop for both models (trading some accuracy for speed). These findings highlight the effectiveness of software-level optimization in distributed learning for medical imaging. This work demonstrates the practical viability of combining architectural design, lightweight obfuscation, and advanced distributed training strategies to accelerate and enhance medical image processing pipelines in real-world clinical and research environments. The full implementation is publicly available at: \url{https://github.com/Suadey/medical-image-denoising-ddp}.

\end{abstract}


\section{Introduction}
Deep learning has significantly advanced medical image analysis, particularly in segmentation and denoising tasks using convolutional neural networks (CNNs) \cite{abdou2022literature}. However, training large-scale models on centralized medical datasets raises certain concerns, such as high computational power demand \cite{lai2015deep} and privacy \cite{aouedi2022handling}. Distributed Deep Learning (DDL) offers a promising alternative, allowing training across multiple GPUs or sites while reducing data centralization.

This study focuses on medical image denoising using chest X-ray images from the NIH ChestX-ray14 dataset \footnote{\href{https://www.kaggle.com/datasets/nih-chest-xrays/data}{NIH Chest X-rays}}. To simulate privacy-preserving data sharing, we apply additive Gaussian noise \cite{russo2003method} as a lightweight obfuscation technique. Noisy images are used as model inputs, while clean images serve as targets, mimicking a real-world scenario where sensitive data is masked before training.

We investigate how distributed training configurations—including single-GPU, multi-GPU (\text{DataParallel}), and optimized \text{DistributedDataParallel (DDP)} with Automatic Mixed Precision (AMP)—affect denoising performance and training efficiency. We use U-Net \cite{ronneberger2015u} and U-Net++\cite{zhou2018unet++}, two widely used CNN architectures in medical imaging, selected for their effectiveness and computational efficiency in image-to-image restoration tasks.

Our contributions are as follows: (1) We develop a distributed deep learning framework for medical image denoising using Gaussian-noised X-ray inputs; (2) We compare U-Net and U-Net++ performance under various GPU configurations; (3) We integrate DDP and AMP for improved training efficiency; and (4) We analyze the viability of noise-based obfuscation as a privacy-preserving mechanism for clinical pipelines.

\section{Literature Review}
Research on distributed deep learning has grown in recent years, with applications ranging from generic image recognition to specialized domains. This section reviews literature that is related to our work. Specifically, we review prior work in three pertinent areas: (a) distributed training to improve efficiency/scalability, (b) privacy-preserving machine learning for sensitive data, and (c) deep learning models for medical image denoising and segmentation.

\subsection{Distributed Deep Learning for Efficiency and Scalability}
Distributed Deep Learning (DDL) has gained momentum for its ability to scale training processes across multiple GPUs or nodes \cite{nurnoby2023distributed, zhang2017poseidon,schaa2009exploring, sun2019mgpusim, zhu2020big}. This has in turn significantly reducing convergence time and enabling the handling of large datasets. Nurnoby et al. \cite{nurnoby2023distributed} showed up to 80\% speedup in convergence time using multi-GPU training for image classification, underscoring the potential of parallelism in accelerating deep learning workflows. In medical imaging, where input sizes are often large and training can be resource-intensive, DDL presents a practical pathway for scaling. Pal et al. \cite{pal2019optimizing} explored both data and model parallelism, finding that hybrid schemes can further improve throughput for deep networks. However, for mid-sized models like U-Net and U-Net++, data parallelism alone is often sufficient and offers lower implementation complexity. Our work leverages PyTorch’s DistributedDataParallel (DDP) with Automatic Mixed Precision (AMP) to maximize performance while maintaining training stability.

\subsection{Privacy-Preserving Machine Learning}
Medical data is inherently sensitive, requiring robust privacy measures during model training and deployment. Federated Learning (FL) frameworks aim to address this by keeping data localized at the source and aggregating only model updates on a central server \cite{nguyen2022federated}. Although FL reduces data exposure, it introduces challenges such as communication overhead, slower convergence, and potential model divergence across sites. Encoding-based methods like Privacy-SF \cite{chen2024privacy} pre-process images into latent representations before training, ensuring raw data is never exposed. Differential Privacy (DP) methods such as DP-SGD \cite{abadi2016deep} further protect individual records by adding calibrated noise to gradients, though often at the cost of model performance. Homomorphic encryption and secure multi-party computation techniques provide strong theoretical guarantees but are computationally expensive and less feasible for large-scale CNNs. We adopt a lightweight privacy mechanism (additive Gaussian noise) as a data obfuscation method that reduces identifiable information without compromising efficiency.

\subsection{Deep Learning Models for Medical Image Denoising and Segmentation}
Denoising is critical in medical imaging for enhancing diagnostic utility, especially when using low-dose acquisition protocols or archival datasets. Convolutional neural networks (CNNs) like U-Net \cite{ronneberger2015u} and its variants remain popular due to their strong performance on segmentation and restoration tasks. U-Net++ \cite{zhou2018unet++} introduces nested skip connections and intermediate layers to enable finer reconstruction of structural features, making it especially valuable in high-noise environments. DnCNN \cite{zhang2017poseidon} introduced residual learning for denoising, and DudeNet \cite{sahu2023application} applied dual-branch designs for robust noise removal on chest X-rays. Traditional filters such as BM3D have largely been replced by CNN-based models that better preserve anatomical fidelity and generalize to variable noise conditions. Given their balance of accuracy, efficiency, and memory footprint, U-Net and U-Net++ are well-suited for distributed training pipelines in clinical scenarios. 
Our work builds upon these foundations by integrating DDL and obfuscation within a practical and scalable denoising framework, a combination not extensively covered in prior research, and by examining both performance gains and privacy considerations together.

\section{Methodology}

\begin{figure}[!ht]
	\centering
	\includegraphics[scale=1]{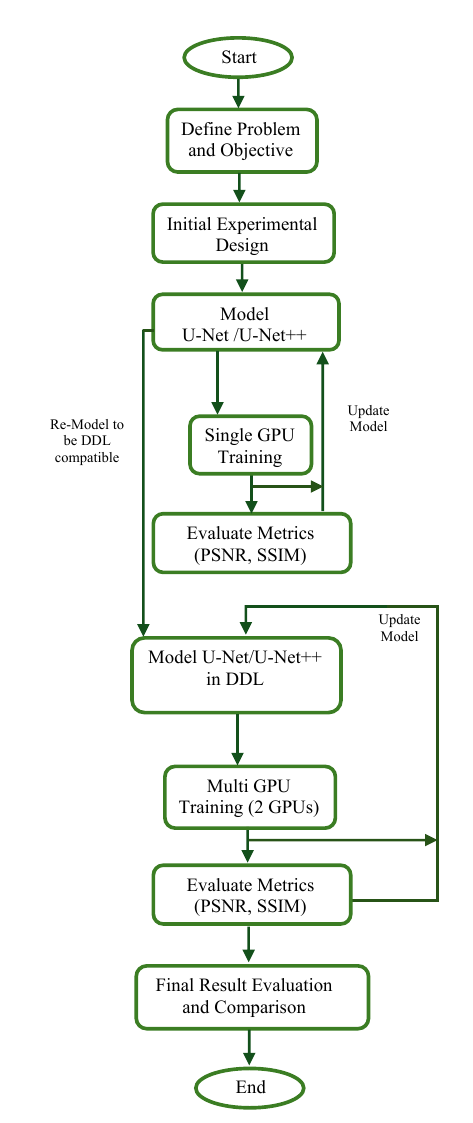}
	\caption{Research Workflow}
	\label{fig:Flow Chart}
\end{figure}

This section describes our distributed deep learning pipeline for medical image denoising, covering data preprocessing, model architectures, training configurations, and evaluation metrics. Our framework is designed to assess the trade-offs between accuracy, speed, and data privacy when using distributed training strategies such as DataParallel and DistributedDataParallel (DDP) with Automatic Mixed Precision (AMP). The research flowchart from the problem definition till the final result evaluation and comparison is given in Figure  \ref{fig:Flow Chart}. 

\begin{figure*}[!ht]
	\centering
	\includegraphics[scale=0.8]{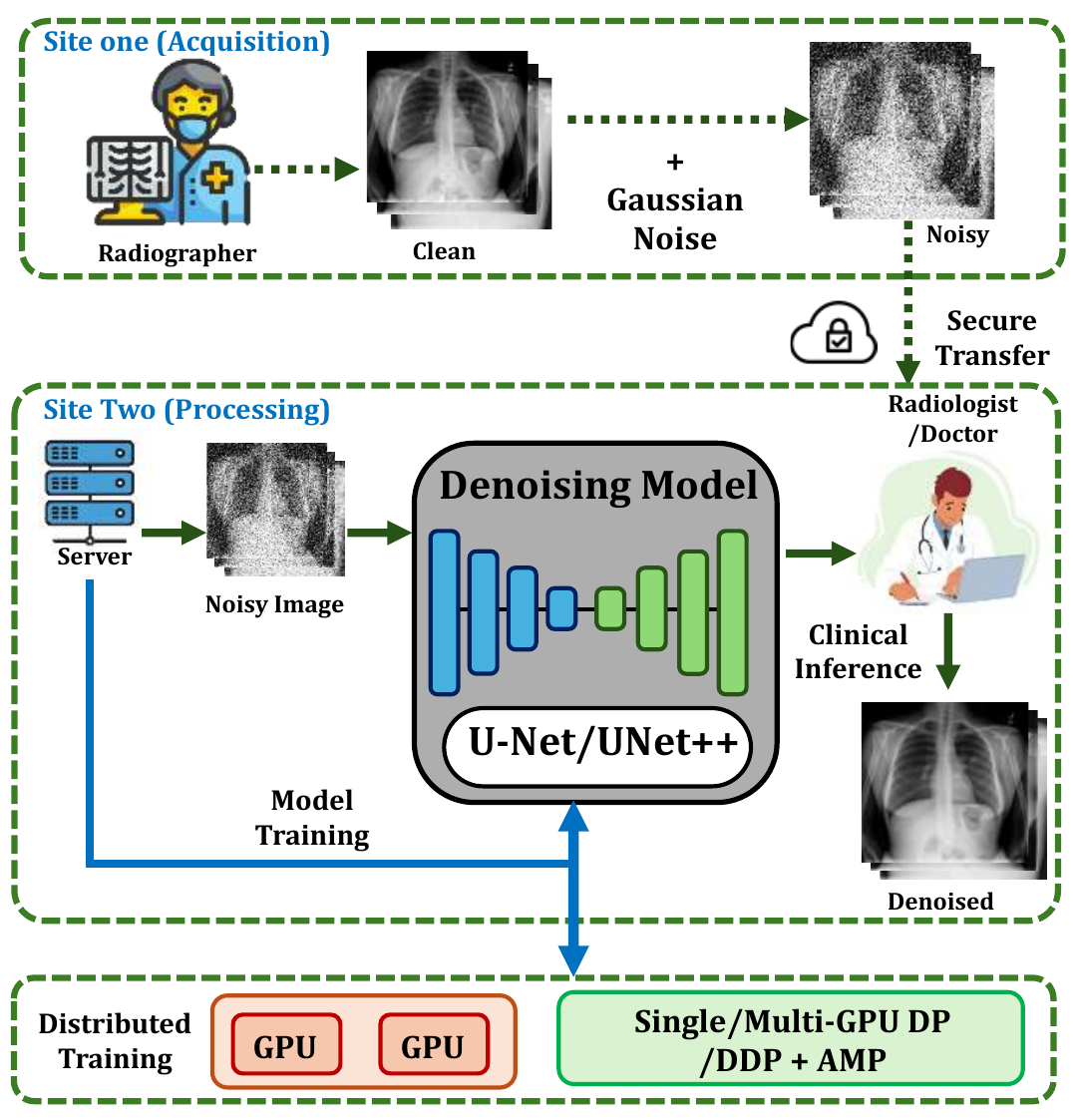}
	\caption{Framework of the proposed distributed denoising pipeline using Gaussian noise-based data obfuscation. At Site One, clean chest X-ray images are acquired and locally obfuscated using additive Gaussian noise. The noisy images are securely transferred to Site Two, where they are processed by a U-Net/UNet++ model. Model training is performed using various GPU configurations—including single GPU, multi-GPU (DataParallel), and DDP with AMP to support scalability and diverse hardware environments. The denoised output is used for clinical interpretation by a radiologist}
	\label{fig:framework}
\end{figure*}

Figure~\ref{fig:framework} presents the overall pipeline. The pipeline is divided into two logical locations: Site One, where chest X-ray images are acquired and noise is injected as a privacy-preserving transformation; and Site Two, where denoising is performed using a U-Net/UNet++ model. A deep learning model (U-Net or U-Net++) is trained on noisy-clean image pairs using multiple GPUs. The resulting denoised images support downstream clinical interpretation. This end-to-end setup enables efficient training and inference under constrained and privacy-conscious conditions.

\subsection{Data Preparation and Noise Obfuscation}

We used 15,000 chest radiographs sampled from the publicly available NIH ChestX-ray14 dataset, originally comprising 112,120 frontal-view X-rays from over 30,000 patients. All images were grayscale and resized from \(1024 \times 1024\) to \(256 \times 256\) to reduce computational cost and enable multi-GPU training with practical batch sizes. This resolution is common in CXR deep learning studies \cite{khater2025attcdcnet,jain2024comparative,nakrani2025advanced} and provides a balance between anatomical detail and tractability \cite{huang2020fusion}.

To simulate a privacy-aware training setup, we applied additive Gaussian noise to the clean images. This obfuscation serves dual purposes: it masks sensitive patient features and creates a supervised denoising task. Gaussian noise with fixed mean 0.1 and standard deviations of 0.1, 0.2, and 0.3 was added to generate 10\%, 20\%, and 30\% noise levels, respectively. By keeping the noise mean fixed at 0.1, we ensure a consistent brightness in the training data.

The noisy images were used as model inputs, and clean images served as targets. Data were split into 7,499 training, 4,949 validation, and 2,551 test pairs (approximately 50/33/17\%). No other augmentations were applied to preserve anatomical orientation. This setup models a realistic scenario in which hospitals share perturbed images while retaining original data locally.

\begin{figure*}[!ht]
	\centering
	\includegraphics[scale=1]{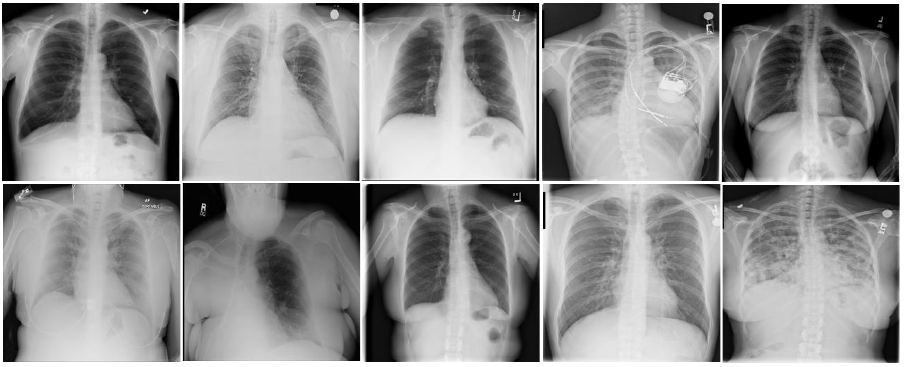}
	\caption{Dataset Sample}
	\label{fig:dataset_sample}
\end{figure*}

\subsection{Denoising Models: U-Net and U-Net++}

We evaluated two encoder–decoder convolutional network architectures (U-Net and U-Net++) for mapping noisy X-ray images to their clean counterparts. Both are widely used in medical imaging for their ability to capture fine anatomical details while remaining computationally tractable.
U-Net consists of a symmetric encoder-decoder structure with skip connections to preserve spatial information. Each encoder block comprises two \(3 \times 3\) convolutions followed by max pooling, and the decoder mirrors this with transposed convolutions and concatenated skip paths.

U-Net++ extends U-Net by incorporating nested skip connections and intermediate convolutional layers across scales. These refinements enhance multi-scale feature aggregation and gradient flow, which can improve restoration quality under high noise levels. Bilinear interpolation is used for upsampling to reduce checkerboard artifacts.

Both models were implemented in PyTorch with a base channel width of 64 and trained using identical protocols. A high-level comparison is summarized in Table~\ref{tab:unet_unetpp_summary}.

\begin{table}[ht]
\caption{\textbf{Summary of U-Net and U-Net++ Architectures Used}}
\label{tab:unet_unetpp_summary}
\setlength{\tabcolsep}{4pt}
\begin{tabular}{|p{25pt}|p{100pt}|p{100pt}|}
\hline
\textbf{Feature} &\textbf{U-Net} & \textbf{U-Net++} \\
\hline
Input & \(1 \times 256 \times 256\) grayscale image & \(1 \times 256 \times 256\) grayscale image \\
\hline
Output & \(1 \times 256 \times 256\) denoised image & \(1 \times 256 \times 256\) denoised image \\
\hline
Depth & 5 levels & 5 levels with nested decoding \\
\hline
Param. & \(\sim 8.6\,\mathrm{M}\) (\(N_c=64\)) & \(\sim 9.2\,\mathrm{M}\) (\texttt{base\_ch}=64) \\
\hline
Key Features & ReLU, BatchNorm, symmetric skip connections, transposed conv upsampling & Dense skip pathways, multi-depth aggregation, multiple output heads, bilinear upsampling\\
\hline
\end{tabular}
\end{table}

\subsection{Distributed Training Setup}

To accelerate training and support scalability, we implemented both \texttt{DataParallel} and \texttt{DistributedDataParallel} (DDP) strategies using PyTorch. \texttt{DataParallel} replicates the model across GPUs and aggregates gradients on the main device (typically GPU 0), while DDP launches one process per GPU with local gradient computation and synchronized updates, leading to better performance and scaling efficiency.

For additional speedup and reduced memory usage, we integrated Automatic Mixed Precision (AMP) via \texttt{torch.cuda.amp}. AMP dynamically casts computations to FP16 where safe, leveraging Tensor Cores on modern GPUs. Gradient scaling was used to maintain numerical stability.

All training was conducted on a workstation with two NVIDIA RTX A4500 GPUs (20~GB memory each), CUDA 12.2, and PyTorch 1.X. DDP+AMP was the fastest and most stable configuration, as detailed in Section~\ref{results}. The details of the training framework and hardware configuration used is presented in Table \ref{tab:training_config} below. 

\begin{table}[ht]
\centering
\caption{\textbf{Training Framework and Hardware Configuration}}
\label{tab:training_config}
\setlength{\tabcolsep}{4pt}
\begin{tabular}{|p{50pt}|p{150pt}|}
\hline
\textbf{Attribute} & \textbf{Details} \\
\hline
Framework & PyTorch 1.X with \texttt{nn.DataParallel} \\
\hline
GPUs Used & 2 \(\times\) NVIDIA RTX A4500 (Each with 20470~MiB memory) \\
\hline
CUDA Version & 12.2 \\
\hline
Driver Version & 535.171.04 \\
\hline
GPU Utilisation & Managed by PyTorch’s automatic batch splitting and gradient synchronization \\
\hline
\end{tabular}
\end{table}

\subsection{Training Protocol and Implementation Details}

All models were trained using the Adam optimizer \cite{kingma2014adam} with an initial learning rate of \(1 \times 10^{-3}\) and L1 loss, which empirically preserved image texture better than L2. Each model was trained for 50 epochs, with early stopping monitored but not triggered. The best model was selected based on minimum validation loss.

Batch size was set to 16 for single-GPU and 32 for dual-GPU setups (16 per GPU). We kept the learning rate fixed across configurations to enable fair comparison, despite theoretical justifications for scaling it linearly with batch size.

We used PyTorch’s native support for AMP and enabled performance optimizations such as pinned memory, preloading, and non-overlapping data loading with \texttt{DistributedSampler}. All models were implemented in PyTorch 1.X and evaluated using the same test dataset.

\subsection{Evaluation Metrics and Visualization}

We assessed model performance using three complementary metrics: Peak Signal-to-Noise Ratio (PSNR), Structural Similarity Index Measure (SSIM), and Learned Perceptual Image Patch Similarity (LPIPS). PSNR and SSIM  (defined in equation \ref{psnr} and \ref{ssim} respectively) quantify pixel-wise fidelity and structural alignment relative to ground truth, while LPIPS reflects perceptual similarity based on deep feature embeddings.

\begin{equation}
\text{PSNR} = 10 \cdot \log_{10} \left( \frac{MAX_I^2}{\text{MSE}(I, \hat{I})} \right),
\label{psnr}
\end{equation}

\begin{equation}
\text{SSIM}(I, \hat{I}) = \frac{(2\mu_I \mu_{\hat{I}} + C_1)(2\sigma_{I\hat{I}} + C_2)}{(\mu_I^2 + \mu_{\hat{I}}^2 + C_1)(\sigma_I^2 + \sigma_{\hat{I}}^2 + C_2)},
\label{ssim}
\end{equation}

All metrics were averaged over a held-out test set of 2,551 images. We report mean values with 95\% confidence intervals. For qualitative analysis, we selected representative samples showing noisy inputs, denoised outputs (U-Net and U-Net++), and the corresponding ground truths to visually evaluate preservation of anatomical details such as ribs, lung textures, and diaphragm contours.

\section{Results and Discussion}
\label{results}

We evaluated U-Net and U-Net++ across three Gaussian noise levels (10\%, 20\%, and 30\%) and two hardware configurations (single-GPU and dual-GPU). Performance was measured using PSNR, SSIM, and LPIPS on a held-out test set of 2,551 images. We also analyzed training efficiency in terms of wall-clock time and GPU utilization.

\subsection{Quantitative Performance and Speedup}

Table~\ref{tab:psnr_ssim_results} summarizes model performance with 95\% confidence intervals. At 10\% noise, U-Net achieved slightly better PSNR and SSIM than U-Net++, while also producing lower LPIPS values, suggesting superior perceptual fidelity. At higher noise levels (20\% and 30\%), U-Net++ consistently outperformed U-Net in PSNR and SSIM, demonstrating better structural preservation. LPIPS scores became comparable at 30\%, where both models converged in perceptual similarity.

\begin{table*}[ht]
\centering
\caption{Denoising performance (PSNR, SSIM, LPIPS) for U-Net and U-Net++ at different noise levels with 95\% confidence intervals.}
\label{tab:psnr_ssim_results}
\begin{tabular}{|c|p{2.1cm}|p{2.1cm}|p{2.1cm}|p{2.1cm}|p{2.1cm}|p{2.1cm}|}
\hline
\textbf{Noise Level} & \multicolumn{3}{c|}{\textbf{U-Net}} & \multicolumn{3}{c|}{\textbf{U-Net++}} \\
\cline{2-7}
 & \textbf{PSNR (dB)} & \textbf{SSIM} & \textbf{LPIPS} & \textbf{PSNR (dB)} & \textbf{SSIM} & \textbf{LPIPS} \\
\hline
10\% & 34.95 (±0.04) & 0.9168 (±0.0008) & 0.1373 (±0.001) & 34.39 (±0.05) & 0.9123 (±0.0007) & 0.1585 (±0.0011) \\
20\% & 32.26 (±0.04) & 0.8907 (±0.0011) & 0.2010 (±0.0013) & 32.32 (±0.05) & 0.8959 (±0.0010) & 0.2265 (±0.0015) \\
30\% & 30.23 (±0.05) & 0.8746 (±0.0012) & 0.2498 (±0.0016) & 30.76 (±0.05) & 0.8840 (±0.0011) & 0.2479 (±0.0016) \\
\hline
\end{tabular}
\end{table*}

All metric differences were statistically significant (\(p < 10^{-13}\)). These results highlight a trade-off: U-Net offers better perceptual alignment at low noise, whereas U-Net++ preserves structure better under heavier degradation.

\subsection{Training Efficiency and Multi-GPU Acceleration}
\begin{figure}[!ht]
	\centering
	\includegraphics[scale=0.8]{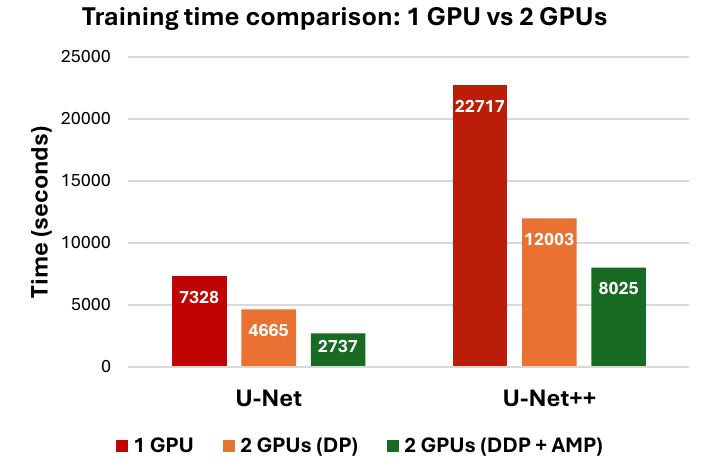}
	\caption{Training time comparison for U-Net and U-Net++ across three configurations: 1 GPU, 2 GPUs with DataParallel, and 2 GPUs with DDP + AMP. The optimized configuration yields the shortest training time for both models.}
	\label{fig:TT_comp}
\end{figure}

We compared training times across three configurations: 1 GPU, 2 GPUs with \texttt{DataParallel}, and 2 GPUs with \texttt{DDP + AMP}. As illustrated in Figure~\ref{fig:TT_comp}, DDP+AMP yielded the largest reduction (over 60\%) in training time outperforming both the single-GPU and \texttt{DataParallel} configurations. Training time was reduced by 36–47\% when moving from 1 GPU to 2 GPUs with \texttt{DataParallel}, and further reduced by over 60\% with the optimized setup. For instance, under 10\% noise, U-Net training time dropped from 7328.2 seconds (1 GPU) to 4665.2 seconds (2 GPUs, \texttt{DataParallel}), and further to just 2737 seconds using DDP + AMP. Similar trends was observed for U-Net++.


Table~\ref{tab:quant_results} shows full results across all noise levels and hardware configurations. U-Net++ have less performance reduction under dual-GPU settings with respect to the performance in single-GPU setup, though it remains more computationally demanding.

\begin{table}[ht]
\centering
\caption{PSNR (dB), SSIM, LPIPS, and training time (in seconds) for U-Net and U-Net++ under varying noise levels and GPU configurations.}
\label{tab:quant_results}
\begin{tabular}{
    |p{18pt}   
    |p{30pt}   
    |p{15pt}   
    |p{22pt}   
    |p{22pt}   
    |p{22pt}   
    |p{30pt}|} 
\hline
\textbf{Noise} & \textbf{Model} & \textbf{GPUs} & \textbf{PSNR (dB)} & \textbf{SSIM} & \textbf{LPIPS} & \textbf{Training Time (s)} \\
\hline
10\% & U-Net     & 1 & 34.952 & 0.9168 & 0.1373 & 7328.2 \\
10\% & U-Net++   & 1 & 34.387 & 0.9123 & 0.1585 & 22717.1 \\
10\% & U-Net     & 2 & 34.034 & 0.9060 & 0.1582 & 4665.2 \\
10\% & U-Net++   & 2 & 34.089 & 0.9083 & 0.1564 & 12003.0 \\
\hline
20\% & U-Net     & 1 & 32.255 & 0.8907 & 0.2011 & 7326.1 \\
20\% & U-Net++   & 1 & 32.315 & 0.8959 & 0.2265 & 22266.0 \\
20\% & U-Net     & 2 & 32.157 & 0.8906 & 0.2143 & 4648.1 \\
20\% & U-Net++   & 2 & 32.151 & 0.8886 & 0.2000 & 11998.5 \\
\hline
30\% & U-Net     & 1 & 30.225 & 0.8746 & 0.2498 & 7322.0 \\
30\% & U-Net++   & 1 & 30.756 & 0.8840 & 0.2479 & 22343.5 \\
30\% & U-Net     & 2 & 30.481 & 0.8757 & 0.2661 & 4648.2 \\
30\% & U-Net++   & 2 & 30.691 & 0.8830 & 0.2638 & 12015.6 \\
\hline
\end{tabular}
\end{table}

\subsection{Optimized Setup: DDP + AMP}
Table~\ref{tab:ddp_summary} summarizes the performance of our optimized DDP+AMP implementation compared to baseline single- and dual-GPU training under 10\% Gaussian noise. Using DDP+AMP led to major speedups: U-Net training time dropped from 7328 to 2737 seconds, and U-Net++ from 22717 to 8025 seconds.

\begin{table*}[ht]
\centering
\caption{Comparison of Denoising Performance, Training Time (TT) in seconds, and Percentage Time Savings (TS) for U-Net and U-Net++ Models under 10\% Gaussian Noise Using Different GPU Configurations}
\label{tab:ddp_summary}
\begin{tabular}{|c|l|c|c|c|c|c|}
\hline
\textbf{Model} & \textbf{Setup} & \textbf{PSNR (dB)} & \textbf{SSIM} & \textbf{LPIPS} & \textbf{TT (s)} & \textbf{TS (\%)} \\
\hline
U-Net   & 1 GPU              & 34.952 & 0.9168 & 0.1373 & 7328  & 0.00\% \\
U-Net   & 2 GPUs (DP)        & 34.034 & 0.9060 & 0.1582 & 4665  & 36.34\% \\
U-Net   & 2 GPUs (DDP + AMP) & \textbf{34.483} & \textbf{0.9067} & \textbf{0.1562} & \textbf{2737}  & \textbf{62.65\%} \\
\hline
U-Net++ & 1 GPU              & 34.387 & 0.9123 & 0.1585 & 22717 & 0.00\% \\
U-Net++ & 2 GPUs (DP)        & 34.089 & 0.9083 & 0.1564 & 12003 & 47.16\% \\
U-Net++ & 2 GPUs (DDP + AMP) & \textbf{33.416} & \textbf{0.8927} & \textbf{0.2175} & \textbf{8025}  & \textbf{64.69\%} \\
\hline
\end{tabular}
\end{table*}

U-Net’s performance held steady across all setups, while U-Net++ saw mild degradation under DDP+AMP. This suggests deeper models may require finer hyperparameter tuning in distributed training.

\subsection{Visual Results and Structural Fidelity}

To qualitatively assess image reconstruction fidelity, we compare denoising outputs at 10\%, 20\%, and 30\% Gaussian noise levels using representative test samples. Each comparison (Done under 1-GPU configurations) includes the noisy input, U-Net output, U-Net++ output, and the clean ground truth. 

\begin{figure*}[!ht]
	\centering
	\includegraphics[scale=0.8]{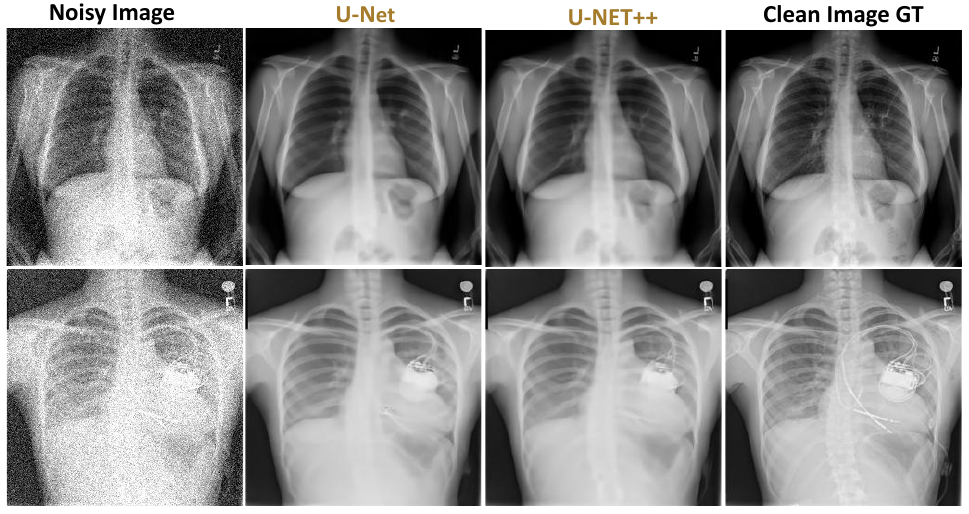}
	\caption{Visual comparison of noisy input, U-Net and U-Net++ denoised outputs, and ground truth (10\% noise levels).}
	\label{fig:qual_10}
\end{figure*}

\begin{figure*}[!ht]
	\centering
	\includegraphics[scale=0.8]{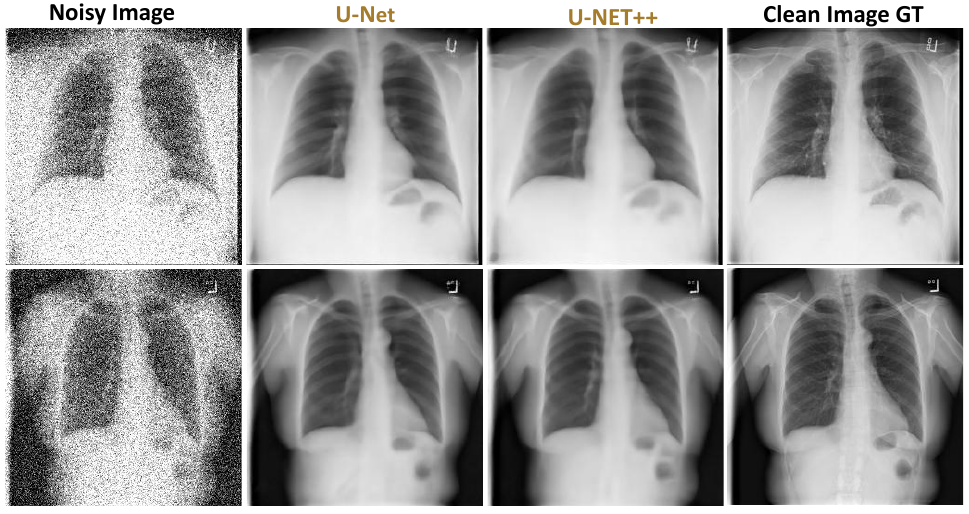}
	\caption{Visual comparison of noisy input, U-Net and U-Net++ denoised outputs, and ground truth (20\% noise levels).}
	\label{fig:qual_20}
\end{figure*}

\begin{figure*}[!ht]
	\centering
	\includegraphics[scale=0.8]{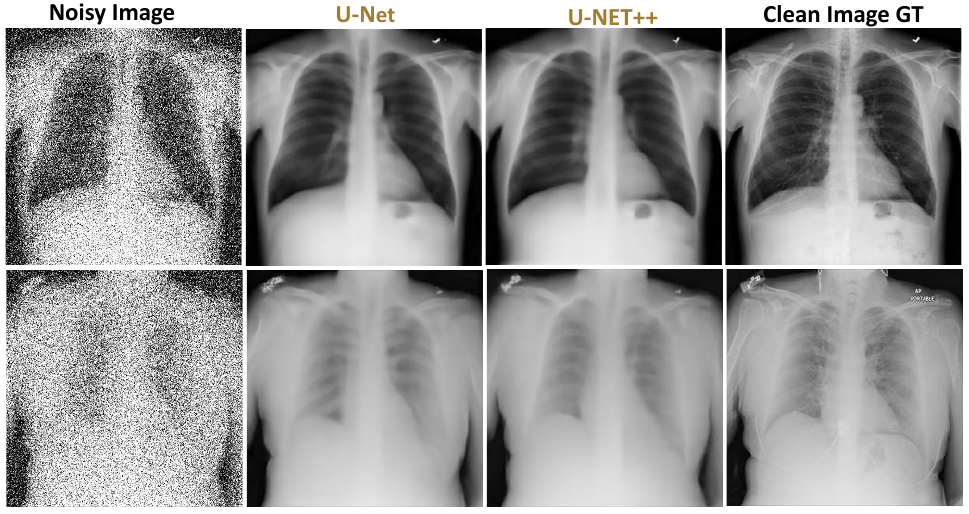}
	\caption{Visual comparison of noisy input, U-Net and U-Net++ denoised outputs, and ground truth (30\% noise levels).}
	\label{fig:qual_30}
\end{figure*}

Figures~\ref{fig:qual_10},~\ref{fig:qual_20}, and~\ref{fig:qual_30} illustrate results at 10\%, 20\%, and 30\% noise levels, respectively.

As shown in Figure~\ref{fig:qual_10}, U-Net produces relatively clean reconstructions, though some residual speckle noise remains. U-Net++, by contrast, yields sharper edge definitions and restores anatomical continuity more faithfully. Structures such as ribs, clavicles, and lung boundaries exhibit enhanced contrast. In particular, the apical regions show smoother transitions and more precise rib contours. The visual SSIM appears higher for U-Net++, with better preservation of brightness consistency and edge contrast.

At 20\% noise (Figure~\ref{fig:qual_20}), U-Net begins to blur finer anatomical details. Mediastinal contours and soft gradients become less distinct. U-Net++, however, maintains better spatial consistency across lung lobes and diaphragm curvature. Noise artifacts that appear in the lower lobes under U-Net are largely suppressed in U-Net++, which retains soft tissue definition. 

At 30\% noise (Figure~\ref{fig:qual_30}), U-Net visibly struggles. Texture smearing and structural loss become prominent, with ambiguous regions such as the vertebral shadow and costophrenic angle. U-Net++ remains relatively robust, with only modest blurring and clear preservation of chest wall and lung structures. Despite strong noise perturbation, U-Net++ maintains perceptual realism and continuity in critical anatomical areas.

\subsection{Comparison with Prior Methods}

We compare our results with other works that employed Guasian Noise for image denoising and recorded the results as given in Table \ref{tab:denoising_results_comp}.

\begin{table}[ht]
\centering
\caption{Denoising performance comparison (PSNR/SSIM) across methods and Gaussian noise levels.}
\label{tab:denoising_results_comp}
\begin{tabular}{|p{60pt}|p{20pt}|p{40pt}|p{20pt}|p{40pt}|}
\hline
\textbf{Method} & \textbf{Noise Level} & \textbf{PSNR/SSIM} & \textbf{Noise Level} & \textbf{PSNR/SSIM} \\
\hline
\textit{OURS (U-Net++)} & 20\% & 32.32/0.8959 & 30\% & 30.76/0.8840 \\
\textit{OURS (U-Net)} & 20\% & 32.26/0.8907 & 30\% & 30.23/0.8746 \\
\textit{BM3D \cite{burger2012image}}           & 15\% & 31.08/0.8722 & 25\% & 28.57/0.8017 \\
\textit{TNDR \cite{chen2016trainable} }          & 15\% & 31.42/0.8826 & 25\% & 28.92/0.8157 \\
\textit{DnCNN-3\cite{zhang2017beyond}}        & 15\% & 31.46/0.8826 & 25\% & 29.02/0.8190 \\
\hline
\end{tabular}
\end{table}

Compared to several established denoising methods, our \textit{U-Net++} approach demonstrates superior performance, particularly under higher noise levels. While prior models such as \textit{BM3D}, \textit{TNDR}, and \textit{DnCNN-3} report PSNR values around 31.1--31.5\,dB and SSIM scores of 0.87--0.88 under 15\% Gaussian noise, our \textit{U-Net++} model achieves a higher PSNR of 32.32\,dB and SSIM of 0.8959 at 20\% noise, indicating better resilience to increased corruption. The performance gap becomes even more pronounced at heavier noise levels: at 25--30\% noise, \textit{U-Net++} achieves 30.76\,dB / 0.8840 SSIM, outperforming the closest competitor, \textit{DnCNN-3} (29.02\,dB / 0.8190), by more than 1.2\,dB in PSNR and over 5 percentage points in SSIM. These results highlight the robustness of our nested architecture in preserving structural details and recovering image fidelity, even under more challenging noise conditions where traditional and early deep learning models begin to degrade noticeably.

\section{Conclusion and Future Work}

This study highlights the efficacy of distributed deep learning for medical image denoising and explores the trade-offs between model complexity and training efficiency. Through systematic experiments, we demonstrated that U-Net++ consistently delivers better structural fidelity and robustness under medium to high Gaussian noise levels, owing to its nested skip connections and dense multi-scale aggregation. In contrast, U-Net provides faster convergence and competitive results, making it a strong candidate for low-noise or real-time scenarios.

Our implementation of an optimized multi-GPU training pipeline using PyTorch's \texttt{DistributedDataParallel} (DDP) and Automatic Mixed Precision (AMP) significantly reduced training time (up to 64\%) without major loss in denoising quality. These improvements affirm the practical value of software-level acceleration techniques in scaling deep learning workflows for clinical imaging, particularly where time or hardware resources are limited.

While U-Net++ incurs a higher computational cost, its perceptual and structural advantages may justify the expense in diagnostic imaging tasks where detail preservation is critical. Selecting between U-Net and U-Net++ should be guided by the clinical context, noise level, and available compute budget.

Looking forward, we aim to enhance both performance and privacy aspects of our framework. Future directions include integrating attention mechanisms to improve fine-grained reconstruction, exploring perceptual loss functions (e.g., VGG or adversarial losses), and conducting human-in-the-loop evaluations with radiologists. We also plan to extend this work into a federated learning paradigm, enabling decentralized training across multiple institutions without exposing sensitive image data.

In conclusion, our work provides a scalable, privacy-conscious, and efficient approach to medical image denoising. These insights contribute to advancing the deployment of distributed deep learning in real-world medical systems and support the development of secure, generalizable AI solutions in healthcare.


\section*{Acknowledgments}
The authors would like to acknowledge all support provided by King Fahd University of Petroleum and Minerals (KFUPM).

\bibliographystyle{ieeetr}
\bibliography{references}

@inproceedings{abadi2016deep,
  title={Deep learning with differential privacy},
  author={Abadi, Martin and Chu, Andy and Goodfellow, Ian and McMahan, H Brendan and Mironov, Ilya and Talwar, Kunal and Zhang, Li},
  booktitle={Proceedings of the 2016 ACM SIGSAC conference on computer and communications security},
  pages={308--318},
  year={2016}
}

@article{abdou2022literature,
  title={Literature review: Efficient deep neural networks techniques for medical image analysis},
  author={Abdou, Mohamed A},
  journal={Neural Computing and Applications},
  volume={34},
  number={8},
  pages={5791--5812},
  year={2022},
  publisher={Springer}
}

@article{aouedi2022handling,
  title={Handling privacy-sensitive medical data with federated learning: challenges and future directions},
  author={Aouedi, Ons and Sacco, Alessio and Piamrat, Karine and Marchetto, Guillaume},
  journal={IEEE Journal of Biomedical and Health Informatics},
  volume={27},
  number={2},
  pages={790--803},
  year={2022},
  publisher={IEEE}
}

@article{chen2024privacy,
  title={Privacy-SF: An encoding-based privacy-preserving segmentation framework for medical images},
  author={Chen, Lei and Song, Lin and Feng, Haoran and Zeru, Robel T and Chai, Shuang and Zhu, En},
  journal={Image and Vision Computing},
  volume={151},
  pages={105246},
  year={2024},
  publisher={Elsevier}
}

@article{lai2015deep,
  title={Deep learning for medical image segmentation},
  author={Lai, Matthew},
  journal={arXiv preprint arXiv:1505.02000},
  year={2015}
}

@article{nguyen2022federated,
  title={Federated learning for smart healthcare: A survey},
  author={Nguyen, Dinh C and Pham, Quoc-Viet and Pathirana, Pubudu N and Ding, Ming and Seneviratne, Aruna and Lin, Zihuai and Li, Jun and Hwang, Woon Hau},
  journal={ACM Computing Surveys (CSUR)},
  volume={55},
  number={3},
  pages={1--37},
  year={2022},
  publisher={ACM}
}

@article{pal2019optimizing,
  title={Optimizing multi-GPU parallelization strategies for deep learning training},
  author={Pal, Saurabh and Ebrahimi, Ehsan and Zulfiqar, Ahsan and Fu, Yuxin and Zhang, Victor and Migacz, Stan and others},
  journal={IEEE Micro},
  volume={39},
  number={5},
  pages={91--101},
  year={2019},
  publisher={IEEE}
}

@inproceedings{ronneberger2015u,
  title={U-Net: Convolutional networks for biomedical image segmentation},
  author={Ronneberger, Olaf and Fischer, Philipp and Brox, Thomas},
  booktitle={International Conference on Medical image computing and computer-assisted intervention},
  pages={234--241},
  year={2015},
  organization={Springer}
}

@article{russo2003method,
  title={A method for estimation and filtering of Gaussian noise in images},
  author={Russo, Francesco},
  journal={IEEE Transactions on Instrumentation and Measurement},
  volume={52},
  number={4},
  pages={1148--1154},
  year={2003},
  publisher={IEEE}
}

@article{sahu2023application,
  title={An application of deep dual convolutional neural network for enhanced medical image denoising},
  author={Sahu, Abhishek and Rana, KPS and Kumar, Vinod},
  journal={Medical \& Biological Engineering \& Computing},
  volume={61},
  number={5},
  pages={991--1004},
  year={2023},
  publisher={Springer}
}

@inproceedings{schaa2009exploring,
  title={Exploring the multiple-GPU design space},
  author={Schaa, Dana and Kaeli, David},
  booktitle={2009 IEEE International Symposium on Parallel \& Distributed Processing},
  pages={1--12},
  year={2009},
  organization={IEEE}
}

@inproceedings{sun2019mgpusim,
  title={Mgpusim: Enabling multi-gpu performance modeling and optimization},
  author={Sun, Yifan and Baruah, Tanmoy and Mojumder, Saiful Azad and Dong, Sheng and Gong, Xi and Treadway, Sean and Ausavarungnirun, Rachata and Kaeli, David},
  booktitle={Proceedings of the 46th International Symposium on Computer Architecture},
  pages={197--209},
  year={2019},
  organization={ACM}
}

@inproceedings{zhang2017poseidon,
  title={Poseidon: An efficient communication architecture for distributed deep learning on {GPU} clusters},
  author={Zhang, Hang and Zheng, Zhihao and Xu, Sheng and Dai, Wei and Ho, Qirong and Liang, Xiaodan and Wang, Hairong and Xing, Eric P},
  booktitle={USENIX Annual Technical Conference (USENIX ATC)},
  pages={181--193},
  year={2017}
}

@inproceedings{zhou2018unet++,
  title={UNet++: A nested u-net architecture for medical image segmentation},
  author={Zhou, Zongwei and Rahman Siddiquee, Md Mahfuzur and Tajbakhsh, Nima and Liang, Jianming},
  booktitle={International Workshop on Deep Learning in Medical Image Analysis and Multimodal Learning for Clinical Decision Support},
  pages={3--11},
  year={2018},
  organization={Springer}
}

@article{zhu2020big,
  title={Big data image classification based on distributed deep representation learning model},
  author={Zhu, Min and Chen, Qingshan},
  journal={IEEE Access},
  volume={8},
  pages={133890--133904},
  year={2020},
  publisher={IEEE}
}

@inproceedings{nurnoby2023distributed,
  title={Distributed Deep Learning-based Model for Large Image Data Classification},
  author={Nurnoby, M Faisal and Shawarib, Khalid A Abu and Khan, Ayaz Ul Hassan},
  booktitle={Proceedings of the 7th International Conference on Future Networks and Distributed Systems},
  pages={283--291},
  year={2023}
}

@article{kingma2014adam,
  title={Adam: A method for stochastic optimization},
  author={Kingma, Diederik P and Ba, Jimmy},
  journal={arXiv preprint arXiv:1412.6980},
  year={2014}
}

@inproceedings{burger2012image,
  title={Image denoising: Can plain neural networks compete with BM3D?},
  author={Burger, Harold C and Schuler, Christian J and Harmeling, Stefan},
  booktitle={2012 IEEE conference on computer vision and pattern recognition},
  pages={2392--2399},
  year={2012},
  organization={IEEE}
}

@article{chen2016trainable,
  title={Trainable nonlinear reaction diffusion: A flexible framework for fast and effective image restoration},
  author={Chen, Yunjin and Pock, Thomas},
  journal={IEEE transactions on pattern analysis and machine intelligence},
  volume={39},
  number={6},
  pages={1256--1272},
  year={2016},
  publisher={IEEE}
}

@article{zhang2017beyond,
  title={Beyond a gaussian denoiser: Residual learning of deep cnn for image denoising},
  author={Zhang, Kai and Zuo, Wangmeng and Chen, Yunjin and Meng, Deyu and Zhang, Lei},
  journal={IEEE transactions on image processing},
  volume={26},
  number={7},
  pages={3142--3155},
  year={2017},
  publisher={IEEE}
}

@article{huang2020fusion,
  title={Fusion high-resolution network for diagnosing ChestX-ray images},
  author={Huang, Zhiwei and Lin, Jinzhao and Xu, Liming and Wang, Huiqian and Bai, Tong and Pang, Yu and Meen, Teen-Hang},
  journal={Electronics},
  volume={9},
  number={1},
  pages={190},
  year={2020},
  publisher={MDPI}
}

@article{nakrani2025advanced,
  title={Advanced Diagnosis of Cardiac and Respiratory Diseases from Chest X-Ray Imagery Using Deep Learning Ensembles},
  author={Nakrani, Hemal and Shahra, Essa Q and Basurra, Shadi and Mohammad, Rasheed and Vakaj, Edlira and Jabbar, Waheb A},
  journal={Journal of Sensor and Actuator Networks},
  volume={14},
  number={2},
  pages={44},
  year={2025},
  publisher={MDPI}
}

@article{jain2024comparative,
  title={A comparative study of cnn, resnet, and vision transformers for multi-classification of chest diseases},
  author={Jain, Ananya and Bhardwaj, Aviral and Murali, Kaushik and Surani, Isha},
  journal={arXiv preprint arXiv:2406.00237},
  year={2024}
}

@inproceedings{khater2025attcdcnet,
  title={AttCDCNet: Attention-enhanced Chest Disease Classification using X-Ray Images},
  author={Khater, Omar H and Shuaibu, Abdullahi S and Haq, Sami Ul and Siddiqui, Abdul Jabbar},
  booktitle={2025 IEEE 22nd International Multi-Conference on Systems, Signals \& Devices (SSD)},
  pages={891--896},
  year={2025},
  organization={IEEE}
}

\end{document}